\documentclass[letterpaper, 10 pt, conference]{ieeeconf}  

\IEEEoverridecommandlockouts                              

\usepackage{booktabs}
\usepackage{multirow}
\usepackage[table,xcdraw]{xcolor}
\usepackage{caption}
\usepackage{graphics} 
\usepackage{epsfig} 
\usepackage{mathptmx} 
\usepackage{times} 
\usepackage{amsmath} 
\usepackage{amssymb}  
\usepackage{pifont}
\usepackage{subcaption}
\usepackage{adjustbox}
\usepackage{array}
\usepackage{multirow, booktabs}
\usepackage{makecell}
\usepackage[mathscr]{euscript}
\usepackage{float}
\usepackage{gensymb}
\usepackage{mathtools}
\usepackage{booktabs}
\usepackage{tabularx}
\usepackage{caption}
\usepackage{subcaption}
\usepackage{graphicx}
\usepackage[linesnumbered,ruled,vlined]{algorithm2e}
\usepackage{algpseudocode}
\usepackage{xcolor}

\usepackage[bookmarks=true, hidelinks]{hyperref}    
\hypersetup{
     colorlinks = true,
     urlcolor = black
}
\usepackage{algpseudocode}
\usepackage{cite}
\usepackage{booktabs}

\title{\LARGE \bf SplatCtrl: Perception-Action Coupling via Gaussian Scene Representations and Reactive Robot Control}

\author{Siddarth Jain$^{1*}$ and Ho Jin Choi$^{1,2*}$
\thanks{$^{1}$Mitsubishi Electric Research Laboratories (MERL), Cambridge, MA, USA. 
{\tt\small sjain@merl.com }}
\thanks{$^{2}$University of Pennsylvania,
Philadelphia, Pennsylvania, USA. This research was completed during H. Choi’s internship at MERL. {\tt\small cr139139@seas.upenn.com }}%
\thanks{$^{*}$ indicates equal contribution.}%
}

\begin{document}

\maketitle
\thispagestyle{empty}
\pagestyle{empty}

\begin{abstract}
Robotic manipulators excel in structured environments but face substantial challenges in unstructured and dynamic settings. This paper presents SplatCtrl, a unified framework for real-time scene reconstruction and reactive robot motion generation to enable collision-free robotic arm control in previously unseen and continuously changing environments. Building on 3D Gaussian Splatting (3D-GS), we introduce a hybrid voxel-based filtering and dynamic Gaussian relocation strategy that supports efficient scene reconstruction from RGB-D streams while accommodating environmental changes. For safe and reactive control, we further propose a method for deriving continuous signed distance functions from isotropic Gaussians, providing stable and differentiable collision probability estimates that bridge classical distance fields with the modern implicit representation. These continuous distance metrics are incorporated into control barrier functions, resulting in a unified  perception-action coupling framework that supports smooth and reliable real-time motion generation in response to scene changes. Experimental validation in simulation, on physical robot, and within shared human-robot workspace demonstrates the framework’s effectiveness, achieving integrated scene reconstruction and reactive control in uncertain, and dynamic environments.

\end{abstract}

\section{INTRODUCTION}

Robotic manipulators demonstrate high levels of performance in structured environments, such as factory floors, where tasks are repetitive and object locations are predetermined. These environments are engineered, often at substantial cost, to enhance repeatability, ensure operational safety, and minimize uncertainty. In contrast, unstructured real-world settings present significantly greater challenges: they are dynamic and unpredictable, with objects and obstacles that may shift or emerge unexpectedly. Under such conditions, robots must rely on sensory inputs to perceive their surroundings, and adapt their motions in real time. 

A key requirement for robotic manipulation in unstructured settings is the ability to construct reliable representations of the environment. Such representations serve as the foundation for perception, planning, and control, yet their design is constrained by the competing demands of fidelity, efficiency, and adaptability. Classical representations~\cite{Hornung2013OctoMapAE, qi2017pointnet}, such as point clouds have been widely adopted due to their simplicity and compatibility. However, they are inherently sparse and often face challenges with occlusions and maintaining temporal consistency. Voxel grids provide a structured representation that facilitates occupancy reasoning, but they face challenges in balancing resolution with computational efficiency. More recent efforts have explored implicit neural representations, such as Neural Radiance Fields (NeRFs)~\cite{mildenhall2020nerf}, which model geometry and appearance as continuous functions. While these approaches achieve impressive scene representation fidelity, their practical deployment is constrained by substantial computational demands and limited adaptability to dynamic environments. Recently, 3D Gaussian Splatting (3D-GS)~\cite{kerbl2023gaussian} has emerged as a more efficient alternative. However, its advantages are largely confined to high-quality rendering of static scenes and fall short of the physical consistency, real-time responsiveness, and multi-modal integration necessary for robotic applications.

\begin{figure}[t]
    \centering
    \includegraphics[width=0.48\textwidth]{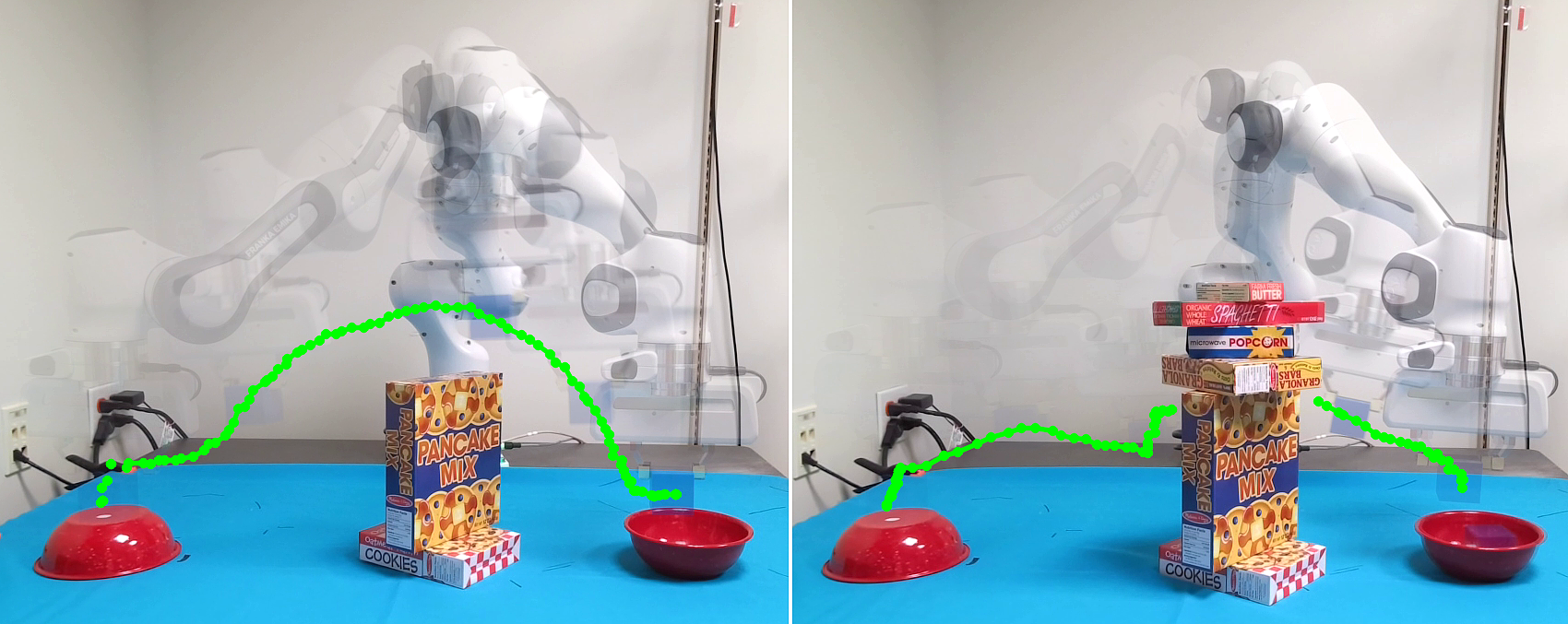}
    \caption{SplatCtrl leverages real-time RGB-D data to reconstruct a scene representation and continuously updates robot's motion accordingly. \textit{Left:} With moderate obstacle height, the robot traverses over the obstacles. \textit{Right:} When obstacle height increased, the updated scene reconstruction triggers an alternative robot motion (shown in {\color{green}green}).}
    \label{fig:main}
    \vspace{-0.7cm}
\end{figure}

Perception-Action Coupling---the continuous interaction between sensory input and motor output, allows robots to adapt their motion based on feedback from the environment~\cite{kappler2018real, huang2023earl}. Conventional methods, which rely on prior knowledge of obstacles for motion generation, struggle to adapt when conditions change. More recent methods leverage motion planning networks~\cite{dalal2024neural} to mimic expert planners, but rely on training with diverse datasets. Achieving greater autonomy demands scene representations that capture both spatial and temporal structure, together with mappings that convert this understanding into reactive motion, without datasets or offline training. This paper presents SplatCtrl (Splat Control), a unified framework  that contributes Gaussian-based scene reconstruction and online reactive robotic control, enabling collision-free robot motion in previously unseen and dynamically evolving environments. The framework builds on the strengths of 3D-GS and extends its applicability to robotics, establishing a principled integration of perception and control through several key contributions: 

\begin{itemize}
    \item \textbf{Extension of 3D Gaussian Splatting (3D-GS):} We introduce a voxel-based filtering with dynamic Gaussian relocation to enable fast, efficient generation and updating of scene reconstructions from RGB-D streams.
    
    \item \textbf{Unified Perception--Action Coupling:} We contribute a method for deriving continuous signed distance functions (SDFs) from isotropic Gaussians, enabling stable and differentiable estimates of collision probability. This approach bridges the gap between discrete distance fields and learning-based representations, while ensuring both computational efficiency and numerical robustness. By integrating these continuous distance representations with control barrier functions, we provide smooth gradient information for reactive robotic control, while preserving accurate distance estimates.

    \item \textbf{Comprehensive Experimental Validation:} We validate our approach through simulation, experiments on a physical robot, and a user study in human–robot workspace, covering a wide range of previously unknown environments with diverse object geometries and spatial arrangements making this the first real-time system capable of achieving full 6-DoF, collision-free robotic control while simultaneously reconstructing scenes from RGB-D streams in previously unknown, and continuously evolving environments.

\end{itemize}

The rest of this article is structured as follows. Section II reviews the related literature. Section III introduces the problem statement and details the proposed approach. Section IV describes the experimental setup and presents the results. Finally, Section V provides the discussion and conclusion.

\section{RELATED WORK}
\label{sec:related}
Scene reconstruction plays a crucial role in robotic perception. Data structures such as point clouds~\cite{zhu2024point}, voxel grids~\cite{zhou2018voxelnet}, and Signed Distance Functions~\cite{breyer2021volumetric} have been used to model environments. Octomap \cite{Hornung2013OctoMapAE}, which models the environment using probabilistic occupancy grids, remains widely used for scene representation and collision detection. Recent advancements in differentiable rendering have significantly enhanced high-fidelity representations. Neural Radiance Fields (NeRF) \cite{mildenhall2020nerf}, model environments as continuous functions, parameterized by neural networks to capture complex geometries and fine details. This approach has found applications in robotics~\cite{wang2024nerfs}, including pose estimation, SLAM, and representation learning. While NeRF excels at generating photorealistic renderings, there is an increasing need for faster methods for latency sensitive applications. 3D Gaussian Splatting (3D-GS) \cite{kerbl2023gaussian} emerged to address this challenge by introducing an advanced explicit scene representation. Recent studies have extended applications of Gaussian Splatting (3D-GS) to robotics~\cite{zhu20243d}. ManiGaussian~\cite{lu2024manigaussian} introduces a framework that parameterize implicit Gaussian points to predict future states and actions. Splat-MOVER~\cite{shorinwa2024splat} builds a modular stack for open-vocabulary manipulation. Robo-GS~\cite{lou2024robo} combines Gaussians, grids, and pixels to reconstruct manipulable robot arms, enabling high-fidelity Real2Sim transfer. SplatSim~\cite{qureshi2024splatsim} jointly reconstructs robots and objects for Sim2Real tasks. Although these methods exploit 3D-GS for precise scene modeling and task-specific planning, they rely on offline predictions and lack the ability to support low-latency, reactive robot motion in uncertain environments. Some approaches~\cite{Michaux2025LetUM, Andreu2025FOCITO}, aim to integrate perception and control more directly. However, they depend on pre-fitted Gaussians and either dense multi-view inputs or expensive scene-specific optimization, which restricts their applicability in dynamic, unstructured settings. Splat-Nav~\cite{Chen2024SplatNavSR} and SAFER-Splat~\cite{chen2024safer} apply online Gaussian updates for robot navigation. Moreover, existing methods exhibit inconsistent rendering and free-space artifacts, which compromise spatial accuracy, a crucial factor in robotic manipulation. Our approach addresses these limitations through voxel-based filtering, dynamic Gaussian relocation, and a reformulation of isotropic Gaussians into a unified distance and collision probability field, thereby enhancing robustness and practical applicability, particularly for robotic manipulators. 

\section{Method}
\label{sec_method}

\begin{figure*}[!t]
    \centering
    \includegraphics[width=1.0\textwidth]{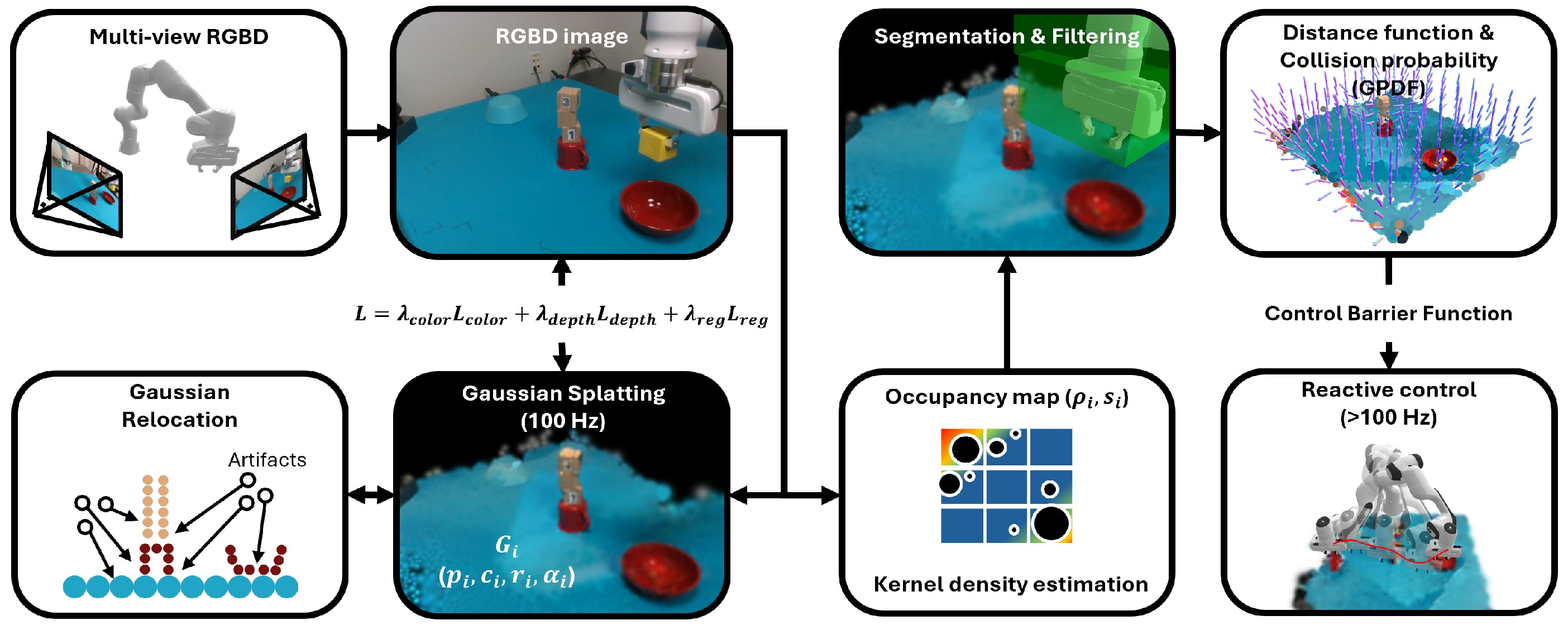}
    \caption{Overview of SplatCtrl, including scene reconstruction via isotropic Gaussian splatting, occupancy-based density estimation and segmentation, and integration with a GPDF to generate a continuous control barrier function.}
    \label{fig:pipeline}
\vspace{-0.6cm}
\end{figure*}

Our method takes RGB-D image streams from multiple cameras as input and builds scene representations within a unified framework (Figure~\ref{fig:pipeline}), enabling seamless coupling between perception and action. The approach is designed to fulfill two key objectives: (1) autonomous, real-time scene reconstruction driven by sensor feedback, and (2) tight integration of perception with control, ensuring that the resulting scene representation directly supports both motion planning and reactive control of robotic arms, even in unfamiliar environments. In the sections that follow, we introduce the framework and explain how it addresses the challenges of applying Gaussian-based scene representations in robotics.

\subsection{Problem Statement}
\label{sec:prob}

We addresses the problem of maintaining an efficient scene representation of a continuously changing workspace $\mathcal{W} \subset \mathbb{R}^3$ for collision-free robotic control in the workspace of the robot. We approximate $\mathcal{W}$ using a set of isotropic Gaussians $\mathcal{G} = \{G_i=(p_i, r_i,...)\}_{i=1}^N$, where $p_i \in \mathbb{R}^3$ and $r_i > 0$ represent the center and radius of each Gaussian, respectively. The key challenge is to construct and efficiently update a proxy Signed Distance Function $\text{SDF}(x): \mathcal{W} \rightarrow \mathbb{R}$ that adapts to changes in $\mathcal{G}$ using Gaussian Process Distance Fields. This $\text{SDF}(x)$ needs to balance accuracy and computational efficiency. The next goal is to enable real-time robot control by using the distance representation and its gradient to formulate a reactive collision-avoidance constraint.

\subsection{3D Gaussian Scene Representation}\label{sec:3d_gaussian_scene_representation}
We represent the scene using isotropic 3D Gaussians \cite{Keetha2023SplaTAMST, Gong2024IsotropicGS} due to their computational simplicity and efficiency in rendering and distance calculations. Each Gaussian, $G_i$, is defined by 8 parameters: position $p_i$ (3), RGB color $c_i$ (3), radius $r_i$ (1), and opacity $\alpha_i$ (1). To simplify calculations, the covariance matrix $\Sigma_i = r_i^2 I$, where $I$ is the 3x3 identity matrix, assumes spherical symmetry. In later stages, we add additional parameters: density $\rho_i$ and segmentation label $s_i$.

This Gaussian representation enables differentiable rendering of color and depth from any viewpoint using RGB-D streams, allowing scene optimization with observed images and poses. In camera coordinates, the Gaussians have covariance $\Sigma'$:
\begin{equation}
    \Sigma' = JW\Sigma W^TJ^T
\end{equation}
where $J$ is the Jacobian of the projective transformation, and $W$ is the view matrix. The pixel color is computed by alpha-blending Gaussians in front-to-back order:
\begin{equation}
    \hat{C} = \Sigma_i c_i \alpha_i \Pi_{j=1}^{i-1} (1-\alpha_i)
\end{equation}
Depth is rendered using the center depth $d_i$ of each Gaussian:
\begin{equation}
    \hat{D} = \Sigma_i d_i \alpha_i \Pi_{j=1}^{i-1} (1-\alpha_i)
\end{equation}

\subsection{Gaussian Splatting for Robotics}\label{sec:dynamic_scene_update}
Multiple cameras and Gaussians enable the original GS to reconstruct a scene, but dynamic environments with changing objects require continuous updates.  

We initialize the Gaussians based on the point cloud data from RGBD cameras, with initial opacity of $\alpha_{\rm init} = 0.5$ \cite{Keetha2023SplaTAMST}. The radius is set based on the pixel size, clamped to a maximum value $r = \min(D_{\rm gt}/f, r_{\rm max})$, where $f$ is the camera's focal length. Gaussian parameters are optimized by rendering images and comparing them to the ground truth \cite{kerbl2023gaussian}. The loss function used for optimization is:
\begin{equation}
    \begin{split}
        & L_{\rm color}=(1-\lambda)L_1(\hat{C}, C_{\rm gt}) + \lambda L_{\rm D-SSIM}(\hat{C}, C_{\rm gt}) \\
        & L_{\rm depth}=L_1(\hat{D}, D_{\rm gt}) \\
        & L_{\rm reg} = \Sigma_i|\alpha_i|\\
        & L=\lambda_{\rm color}L_{\rm color} + \lambda_{\rm depth}L_{\rm depth} + \lambda_{\rm reg}L_{\rm reg}
    \end{split}    
\end{equation}
where $L_1$ is least absolute deviations and $L_{\rm D-SSIM}$ is a loss for difference of structural similarity. Here, $\lambda = 0.2$, $\lambda_{\rm color} = \lambda_{\rm depth} = 1$, and $\lambda_{\rm reg} = 0.02$. We use RGBD images to reduce the reconstruction error for high fidelity. The regularization term, $L_{\rm reg}$, encourages fewer Gaussians. Noise is added after each Adam optimizer update. In dynamic situations, Gaussians need to be added, removed, or relocated beyond the optimization. A mask is first generated to identify where to add new Gaussians \cite{Keetha2023SplaTAMST}:
\begin{equation}
    M_{\rm add} = (D_{\rm gt} < \hat{D})\wedge(L_1(\hat{D}, D_{\rm gt}) > \lambda_{\rm MDE} \rm MDE)
\end{equation}
This mask adds Gaussians where the ground truth depth is closer than the rendered depth, and the depth error exceeds $\lambda_{\rm MDE}$ times the median depth error (MDE). Gaussian Splatting generates artifacts that can potentially cause false positives collision. To mitigate this issue, we apply a minimum opacity threshold of \(\alpha_{\rm min} = 0.1\) to filter out transparent Gaussians. Additionally, we use a GPU-based occupancy grid with voxel sizes at least as large as the maximum Gaussian radius. This enables us to evaluate the occupancy values (from 0 to 1) of neighboring voxels and remove Gaussians located in free (\(<0.3\)) or uncertain (\(<0.7\)) regions. Gaussians outside the robot's reachable workspace are also removed.
We relocate Gaussians to adapt to newly appearing objects using the mask $M_{\rm add}$ and RGB-D image. To maintain computational efficiency, we impose a limit on the maximum number of Gaussians. If the total Gaussian count is below the threshold, new Gaussians are added and dead ones are repositioned. Once the limit is reached, only dead Gaussians are replaced, maintaining a fixed count.

To enable obstacle avoidance and environment interaction, the robot must identify Gaussians representing itself. Segmenting from 2D images misses internal Gaussians, causing false collisions. Instead of complex learning-based methods, we use an occupancy grid and the robot’s axis-aligned bounding box (AABB): voxels intersecting the AABB are found, followed by a fixed-radius search for intersecting Gaussians. We use 2-bit encoding to categorize entities: for example objects are 00, the robot is 01. Voxel grids are updated using atomic OR operations to track which entity occupies each voxel. As the robot moves quickly, Gaussians may leave ``trails" where they were previously, which aren’t updated quickly enough by the Gaussian Splatting. To address this, we add a removal mask for Gaussians previously classified as robot that have become objects.

\subsection{SDF and Collision Probability}\label{sec:proxy_sdf_and_collision_probability}

SDF provides information about whether a point is inside, outside, or on the surface of an object, enabling more nuanced robotic interactions.
Isotropic Gaussians can define a surface boundary by their radii $r_i$, naturally producing an SDF \cite{Jin2024GSPlannerAG}. The SDF at a query point $x$ in 3D space is:
\begin{equation}\label{sdf_min}
    \rm SDF (x) = \min_{i} (\|x-p_i\|_2-r_i)
\end{equation}
Here, $p_i$ and $r_i$ denote the center and radius of the $i$-th Gaussian. This method suffers from discontinuities, non-differentiability, and poor surface interpolation under sparse Gaussians.
To address these issues, we extend the Gaussian process distance field (GPDF) \cite{Gentil2023AccurateGD} from point cloud data to include radii. We use it as a proxy for SDF and we improve its numerical stability and computational efficiency by modeling the occupancy field $o(x)$ as a Gaussian process:
\begin{equation}
    o(x) \sim \mathcal{GP}(0, \mathbf{k}(x, p))
\end{equation}
Here, $x \in \mathbb{R}^3$ is the query point, and $p \in \mathbb{R}^3$ is the center of a Gaussian sphere, and the kernel function $\mathbf{k}(x, p) = \mathbf{k}_d(\mathbf{d}(x,p)) = \exp(-\mathbf{d}(x,p) / l)$ is a Mat\'ern kernel with interpolation parameter $l > 0$ where $\mathbf{d}(x,p)=\|x-p\|_2$. The inferred occupancy $\hat{o}(x)$ is computed as:
\begin{equation}
    \hat{o}(x) = \mathbf{k}(x, \mathbf{P})(\mathbf{K}(\mathbf{P}, \mathbf{P}) + \epsilon I)^{-1} \mathbf{y} > 0
\end{equation}
where $\mathbf{P} = [p_1, p_2, \dots, p_n]^T$ are the sphere centers, $\mathbf{y} = [\mathbf{k}_d(-r_1), \mathbf{k}_d(-r_2), \dots, \mathbf{k}_d(-r_n)]^T$ are the kernel evaluations at negative radii, $\epsilon > 0$ is a small noise for observation. The occupancy describes the point's location relative to the object:
\[
\hat{o}(x) < 1: \text{outside}, \quad
= 1: \text{surface}, \quad
> 1: \text{inside}
\]
Although a Gaussian Process (GP) output may not always be positive despite positive observations $\mathbf{y}$, $\hat{o}(x)$ generally stays positive. This positivity is guaranteed under an approximation introduced later. The SDF is defined as:
\begin{equation}
    \widehat{\rm SDF}(x)=\mathbf{r}(\hat{o}(x))
\end{equation}
where $\mathbf{r}$ is the inverse function of $\mathbf{k}_d$. The gradient of SDF is
\begin{equation}
    \begin{split}
        & \nabla\widehat{\rm SDF}(x)=\frac{\partial r}{\partial \hat{o}} \nabla \hat{o}(x)\\
        & \nabla \hat{o}(x) = \nabla \mathbf{k}(x, \mathbf{P}) (\mathbf{K}(\mathbf{P}, \mathbf{P}) + \epsilon I)^{-1} \mathbf{y}
    \end{split}
\end{equation}
To satisfy the Eikonal equation, the gradient is normalized:
\begin{equation}
    \nabla\widehat{\rm SDF}(x)=\frac{\nabla \hat{o}(x)}{\|\nabla \hat{o}(x)\|_2}  
\end{equation}
The interpolated proxy SDF typically underestimates true surface distance. While sufficient for barrier functions, sphere marching refines the SDF in 3–5 iterations for higher accuracy \cite{choi2024grasping}. Alternatively, kernels such as the Gaussian $\mathbf{k}_d(\mathbf{d}) = \exp(-\mathbf{d}^2 / (2l^2))$ yield sharper distance fields but may cause numerical discontinuities.

\subsubsection{Numerical stability}
\begin{table}[ht]
    \centering
    \caption{Numerical stability analysis across varying values of $l$ for a 2D unit circle point cloud.}
    \begin{tabular}{lccccccc}
    \toprule
    Parameter $l$ & $10^{0}$ & $10^{-2}$ & $10^{-4}$ & $10^{-6}$ & $10^{-8}$ & $10^{-10}$\\
    \midrule
    GPDF w/o Eq~\ref{eq:gpdf_numerical} & \checkmark & \checkmark & $\times$ & $\times$ & $\times$ & $\times$\\
    GPDF w/ Eq~\ref{eq:gpdf_numerical}   & \checkmark & \checkmark & \checkmark & \checkmark & \checkmark & \checkmark \\
    \bottomrule
    \end{tabular}
    \label{tab:gpdf_comparison}
\end{table}
GPDF faces numerical instability from $\mathbf{r}(x) = -l \log(x)$, which diverges for points far from $P$ \cite{goel2024distance} as in Table~\ref{tab:gpdf_comparison}. We mitigate this using the following trick:
\begin{equation}
    \begin{split}
        & \mathbf{M} = [m_1, m_2, \dots, m_n]^T = (\mathbf{K}(\mathbf{P}, \mathbf{P}) + \epsilon I)^{-1} \mathbf{y}\\
        & \hat{o}(x) = \mathbf{k}(x,p_1)m_1 + \mathbf{k}(x,p_2)m_2 + \cdots + \mathbf{k}(x,p_n)m_n \\
        & \widehat{\rm SDF}(x) = -l \log(\mathbf{k}(x,p_1)m_1 + \cdots + \mathbf{k}(x,p_n)m_n) \\
        & = \mathbf{d}(x,p_1) -l \log(m_1  + \cdots + \mathbf{k}_d(\mathbf{d}(x,p_n) - \mathbf{d}(x,p_1))m_n)
    \end{split}
    \label{eq:gpdf_numerical}
\end{equation}
where $\mathbf{d}(x,p_1)$ is the minimum distance between $x$ and all points in $\mathbf{P}$. Although the log argument may not always be positive since $\mathbf{M}$ can contain negative elements, it usually is, and suitable approximations ensure positivity.

\subsubsection{Collision probability}
The variance of the Gaussian process provides a measure of collision probability:
\begin{equation}
    \mathbf{var}(o(x)) = \mathbf{k}(x, x) - \mathbf{k}(x,\mathbf{P})(\mathbf{K}(\mathbf{P},\mathbf{P})+\epsilon I)^{-1}\mathbf{k}(\mathbf{P}, x)
\end{equation}
The collision probability $P(o(x) \geq 1)$ where $o(x)>0$ is:
\begin{equation}
    P(o(x) \geq 1) = \frac{
    1 - \Phi\left(\frac{1-\hat{o}(x)}{\sqrt{\mathbf{var}(o(x))}}\right)
    }{1 - \Phi\left(\frac{0-\hat{o}(x)}{\sqrt{\mathbf{var}(o(x))}}\right)}
\end{equation}
Here, $\Phi$ represents the cumulative distribution function of the standard normal distribution. However, when $x$ is far from $\mathbf{P}$, $\hat{o}(x)$ and $\mathbf{var}(o(x))$ approach 0 and 1, respectively, causing $P(o(x) \geq 1)$ to have a non-zero probability of approximately $(1 - \Phi(1))/(1 - \Phi(0)) \approx 0.3174$. This is because $\mathbf{var}(o(x))$ does not account for the scale of $y$. To ensure the probability decreases to near zero as $x$ moves away from $\mathbf{P}$, we divide the variance by a constant factor (e.g., we use 9).

\subsubsection{Approximation}
To reduce the complexity of matrix inversion in Gaussian processes, we apply a mass-lumped matrix approximation \cite{Sellan2022StochasticPS}. We approximate the kernel matrix $\mathcal{K}(\mathcal{P}, \mathcal{P})$ by its row sums, forming a diagonal matrix $\mathcal{D}(\mathcal{P}, \mathcal{P})$:
\begin{equation}
    \begin{split}
        & \mathbf{K}(\mathbf{P}, \mathbf{P}) \approx \mathbf{D}(\mathbf{P}, \mathbf{P}) \\
        & = \rm diag \left(\left[\sum_{i=1}^n \mathbf{k}(p_1, p_i),\dots, \sum_{i=1}^n \mathbf{k}(p_n, p_i)\right]^T\right)
    \end{split}
\end{equation}
This approximation performs kernel regression on the Gaussian process implicit features $\mathbf{F}$, producing new features $\mathbf{F}'$.
\begin{equation}
    \begin{split}
        & \mathbf{K}(\mathbf{P}, \mathbf{P})\mathbf{F} = \mathbf{y} ,\quad \mathbf{D}(\mathbf{P}, \mathbf{P})\mathbf{F}' = \mathbf{y} \\
        & \Rightarrow \mathbf{F}' = \mathbf{D}^{-1}\mathbf{K}\mathbf{F} =
        \begin{bmatrix}
            \frac{\mathbf{k}(p_1, p_1)}{\sum_{i=1}^n \mathbf{k}(p_1, p_i)} & \cdots & \frac{\mathbf{k}(p_1, p_n)}{\sum_{i=1}^n \mathbf{k}(p_1, p_i)} \\
            \vdots & \ddots & \vdots \\
            \frac{\mathbf{k}(p_n, p_1)}{\sum_{i=1}^n \mathbf{k}(p_n, p_i)} & \cdots & \frac{\mathbf{k}(p_n, p_n)}{\sum_{i=1}^n \mathbf{k}(p_n, p_i)}
        \end{bmatrix} \mathbf{F}
    \end{split}
\end{equation}
This reduces model complexity to $O(n^2)$ and ensures $\hat{o} > 0$. The diagonal terms approximate kernel density, e.g., $\sum_{i=1}^n \mathbf{k}(p_1, p_i)$ estimates the density at $p_1$. To simplify further, we approximate density via a uniform voxel grid: rasterize $\mathbf{P}$ using trilinear interpolation, apply a GPDF-based kernel convolution, then interpolate back to get densities $\rho_i$. This reduces complexity to linear in number of points and voxels.

\subsection{Collision Avoidance with Reactive Control }\label{sec:collision_avoidance_with_qp_ik}
We use SDF as a control barrier function for a Quadratic Programming Inverse Kinematics (QP-IK) controller \cite{Salehian2018AUF, Zhang2024ConstrainedPI}. This method enables reactive and collision-free control in dynamic environments. The QP-IK controller is:
\begin{equation}
\begin{split}
\min_{\dot q, \mathbf{\dot x}} \quad & (\mathbf{\dot x}_{\rm ref}- \mathbf{\dot x})^T Q(\mathbf{\dot x}_{\rm ref}- \mathbf{\dot x})+ \delta_{\mathbf{x}}^T R \delta_{\mathbf{x}} +\delta_{\rm ext}^T S \delta_{\rm ext}  \\
\text{s.t.}\quad & \mathbf{\dot x} = J(q) \dot q + \delta_{\mathbf{x}}\quad\quad\quad\text{(Forward kinematics)} \\
&-\dot q \ge - (q_{\rm max} - q),\quad\text{(Joint upper limit)}\\
&\dot q \ge - (q - q_{\rm min}),\quad\quad\,\,\text{(Joint lower limit)}\\
& -\dot q_{\rm \max} \leq \dot q \leq \dot q_{\rm \max}, \quad\text{(Joint velocity limit)}\\
&\nabla_q d_{\rm ext}(q) \dot q \ge -(d_{\rm ext}(q) - \epsilon) + \delta_{\rm ext}, \\
& \text{(External collision constraint)} \\
&\nabla_q d_{\rm self}(q) \dot q \ge -(d_{\rm self}(q) - \epsilon) \\
& \text{(Self collision constraint)} \\
\end{split}
\end{equation}
Here, $\mathbf{\dot{x}}_{\rm ref} \in \mathbb{R}^6$ is the desired end-effector velocity, $J(q) \in \mathbb{R}^{6 \times 7}$ the Jacobian, $\delta_{\mathbf{x}} \in \mathbb{R}^6$ and $\delta_{\rm ext} \in \mathbb{R}^7$ are slack variables, and $Q, R, S \succ 0$ are weights. Joint limits are $q_{\min}, q_{\max}$, and joint velocity limit is $\dot{q}_{\max}$. Collision distances and gradients are $d_{\rm ext}(q), \nabla_q d_{\rm ext}(q)$ and $d_{\rm self}(q), \nabla_q d_{\rm self}(q)$. The cost includes tracking error, slack penalties for kinematics, and soft collision avoidance, with optional terms for velocity smoothness and home pose bias. Collisions are modeled using spheres \cite{Sundaralingam2023CuRoboPC}, and QP solves at $>100$ Hz using CVXPY~\cite{diamond2016cvxpy}'s base solver with state and Gaussian updates.

\section{Experiments \& Results}
\label{sec:experiment}

We validate our framework through simulation, physical robot experiments, and a user study in a shared human–robot workspace, spanning diverse and previously unseen environments with varied object geometries and arrangements.

\subsection{Scene Reconstruction Quality}

\begin{figure}[h]
\begin{subfigure}{0.32\linewidth}
\includegraphics[width=\linewidth]{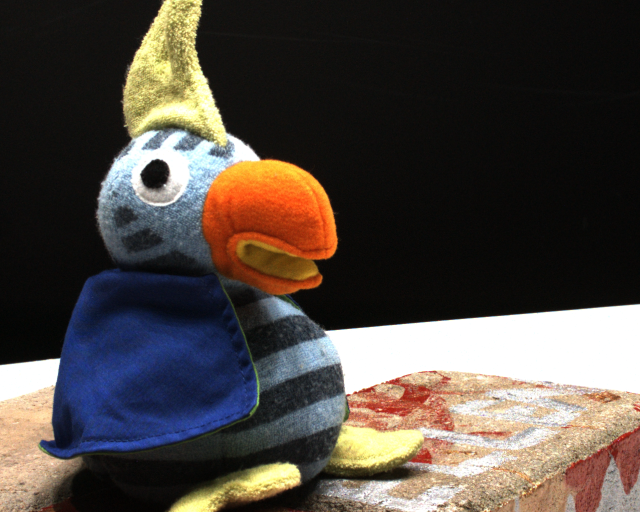} 
\caption{Dataset image}
\end{subfigure}
\hfill
\begin{subfigure}{0.32\linewidth}
\includegraphics[width=\linewidth]{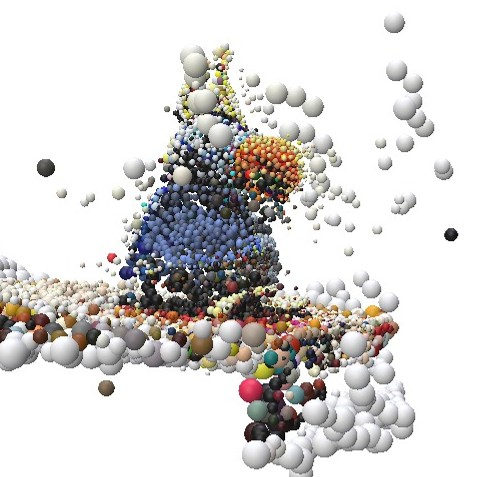} 
\caption{3D-GS~\cite{kerbl2023gaussian}}
\end{subfigure}
\hfill
\begin{subfigure}{0.32\linewidth}
\includegraphics[width=\linewidth]{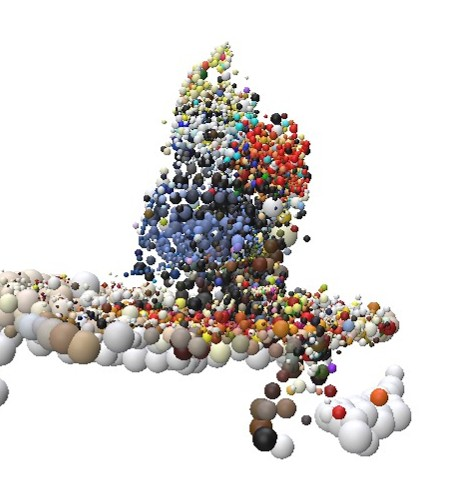} 
\caption{Ours}
\end{subfigure}
\caption{The qualitative comparison of our method with
3D-GS on the DTU MVS dataset~\cite{jensen2014large}.}
\label{fig:dtu_reconstruction}
\vspace{-0.2cm}
\end{figure}

We evaluate scene reconstruction quality using the DTU MVS dataset~\cite{jensen2014large}. Since DTU lacks native depth images, we use depth maps from~\cite{yao2018mvsnet} to synthesize RGB-D data. We benchmark our method against 3D-GS~\cite{kerbl2023gaussian}. Both methods are initialized with $10{,}000$ Gaussians and fitted using identical loss functions. While 3D-GS dynamically increases its count of Gaussians ($50{,}000$–$100{,}000$), 
our method maintain a fixed budget of Gaussians. 
We report quantitative results using Peak Signal-to-Noise Ratio (PSNR, higher values ar better). The baseline 3D-GS method achieves a PSNR of $23.27$, whereas our proposed approach attains a PSNR of $23.33$, with the improvement primarily driven by our Gaussian relocation algorithm, which optimizes Gaussian placement in critical regions~\cite{Kheradmand20243DGS}. This reflects a modest improvement in reconstruction quality, accompanied by enhanced geometric accuracy and reduced artifacts (see Figure~\ref{fig:dtu_reconstruction}).

\begin{figure}[h]
    \centering
    \includegraphics[width=0.48\textwidth]{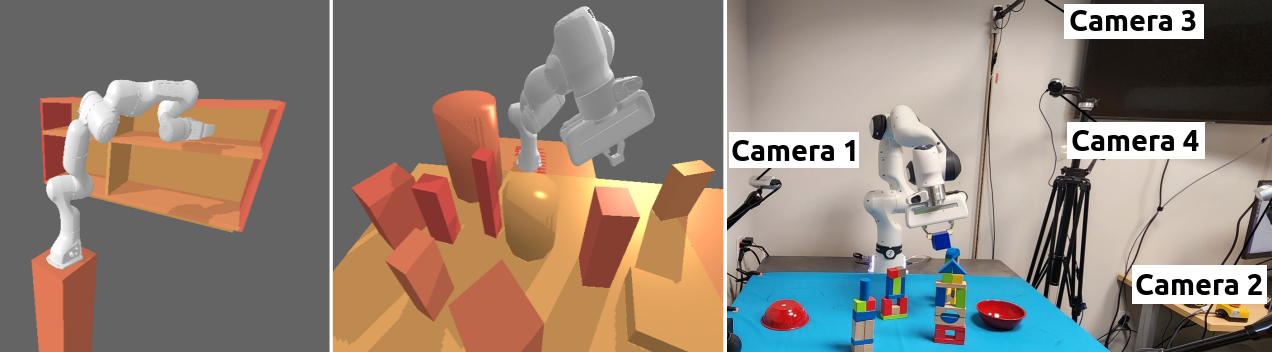}
    \caption{\textit{Left and Middle:} Illustrations of the Cubby and Tabletop simulation environments \cite{Fishman2022MotionPN}. \textit{Right:} Experimental setup for real-robot tasks. Three Intel RealSense D435 cameras (Cameras 1–3) provide multi-view RGB-D observations for scene reconstruction. For the human–robot study, an additional Intel L515 camera (Camera 4) is included.}
    \label{fig:environments}
\vspace{-0.2cm}
\end{figure}

\begin{figure*}[t]
    \centering
    \includegraphics[width=1.0\textwidth]{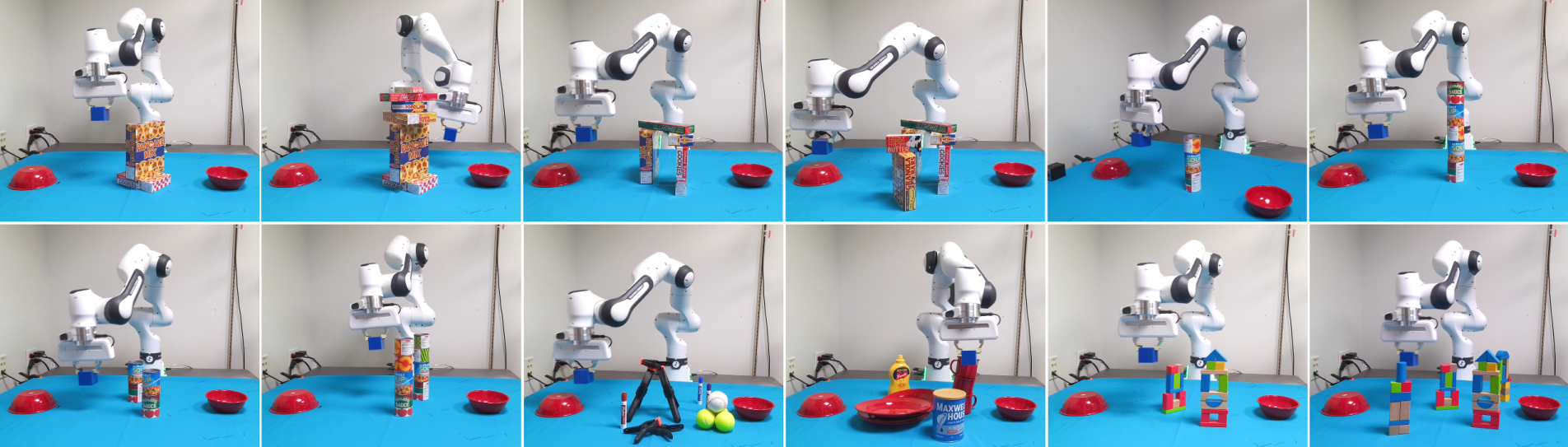}
    \caption{Evaluation in twelve unseen environments containing obstacles that vary in geometry, layout, color, height, and distribution. In each scenario, real-time scene reconstruction and collision-free motion generation are performed as the robot transports a blue cube into a bowl on the right while avoiding collisions.}
    \label{fig:real_envs}
\vspace{-0.4cm}
\end{figure*}

\subsection{Simulation Robot Experiments}  

Reconstruction quality primarily measures scene fidelity, but it does not reflect how useful the model is for downstream robotic tasks. To evaluate utility, we tested SplatCtrl in two simulation environments~\cite{Fishman2022MotionPN}, demonstrating its ability to work with a sampling-based motion planner (MP). Across 942 trials in previously unseen environments (Figure~\ref{fig:environments}, left and middle), the system was provided only with start and goal configurations along with simulated camera RGB-D streams. A sampling-based motion planner (BiRRT) then used the reconstructed scene from our method to generate and execute collision-free trajectories. We benchmarked against 3D-GS~\cite{kerbl2023gaussian} as an alternative scene representation. For collision checking, BiRRT employed the GPDF with $2$ $cm$ voxels, where $\text{GPDF} > 0$ indicated free space. A trial was considered successful if the robot reached the target pose within $1^\circ$ of rotation and $3$ $cm$ of translation without collisions. To further analyze the role of multi-view reconstruction, we conducted an ablation study by varying the number of RGB-D cameras (2–6) available for reconstruction in both our method and the baseline. 

\begin{table}[h]
    \centering
    \caption{Success rates (\%) of motion planning in simulation with reconstructed scenes.}    
    \footnotesize
    \begin{tabularx}{\linewidth}{Xccc}
    \toprule
    Number of Cameras & 2 & 4 & 6 \\
    \midrule
    \multicolumn{4}{l}{\textbf{Tabletop Environment} (546 trials)} \\
    3D-GS w/ MP (\%) & 50.59 & 78.43 & 33.88 \\
    SplatCtrl w/ MP (\%) & \textbf{95.28} & \textbf{92.87} & \textbf{98.17} \\
    \midrule
    \multicolumn{4}{l}{\textbf{Cubby Environment} (174 trials)} \\
    3D-GS w/ MP (\%) & 6.45 & 1.24 & 0.00 \\
    SplatCtrl w/ MP (\%) & \textbf{95.48} & \textbf{87.58} & \textbf{88.51} \\
    \midrule
    \multicolumn{4}{l}{\textbf{Merged Cubby Environment} (222 trials)} \\
    3D-GS w/ MP (\%) & 8.04 & 4.19 & 0.45 \\
    SplatCtrl w/ MP (\%) & \textbf{93.97} & \textbf{94.42} & \textbf{96.85} \\
    \midrule
    \multicolumn{4}{l}{\textbf{Average Across Environments} (942 trials)} \\
    3D-GS w/ MP (\%) & 32.83 & 48.13 & 19.75 \\
    SplatCtrl w/ MP (\%) & \textbf{95.01} & \textbf{92.32} & \textbf{96.07} \\
    \bottomrule
    \end{tabularx}
    \label{table:success_rate_sim}
\vspace{-0.2cm}
\end{table}

Table~\ref{table:success_rate_sim} reports success rates for the simulation experiments. SplatCtrl consistently outperforms the baseline, achieving higher reliability with fewer cameras and demonstrating greater robustness and efficiency. In contrast, the baseline often introduces artifacts that manifest as false collisions, particularly at the start and end-effector positions. While increasing the camera viewpoints can enhance success rates, it also raises the computational demand for Gaussian fitting and consequently, with a fixed iteration budget, additional views may introduce artifacts that can degrade planning performance. Our approach produces more stable scene representations while maintaining consistent performance and computational efficiency, achieving approximately 240 Hz for single-view iterations with a limited Gaussian budget.

\begin{figure}[!t]
    \begin{subfigure}{0.32\linewidth}
    \includegraphics[width=1.0\linewidth]{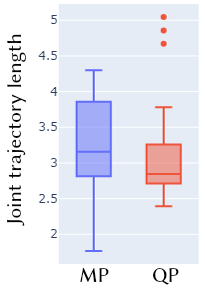} 
    \caption{Joint length}
    \label{fig:real_envs_joint}
    \end{subfigure}
    \begin{subfigure}{0.32\linewidth}
    \includegraphics[width=1.0\linewidth]{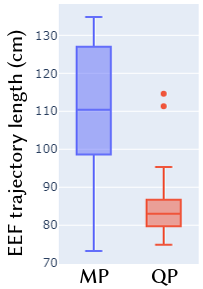}
    \caption{EEF length}
    \label{fig:real_envs_ee}
    \end{subfigure}
    \begin{subfigure}{0.32\linewidth}
    \includegraphics[width=1.0\linewidth]{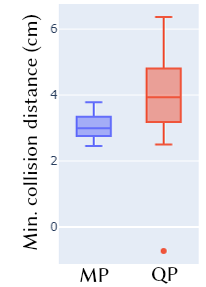}
    \caption{Min distance}
    \label{fig:real_envs_min}
    \end{subfigure}
    \caption{Comparison of perception–action coupling using SplatCtrl with motion planning (MP) and reactive control (QP-IK) across twelve real-world environments.}
    \label{fig:real_results}
\vspace{-0.4cm}
\end{figure}

\subsection{Real Robot Experiments}

We evaluate our framework on a physical robotic platform (Franka Emika Panda FR3), assessing its ability to execute full 6-DoF, collision-free motions while reconstructing scenes in real time from RGB-D streams (Figure~\ref{fig:environments}, right). Experiments are conducted across twelve previously unseen environments that encompass a diverse range of obstacles that vary in geometry, spatial configuration, color, and distribution (Figure \ref{fig:real_envs}). In each environment, the robot is tasked with transporting a blue cube into a bowl positioned on the right, while avoiding collisions. To highlight the modularity, we integrated and evaluated two distinct motion generation pipelines representing complementary control paradigms. The first is our proposed reactive controller (QP-IK), which continuously adapts to the reconstructed scene. The second is a deliberative, sampling-based planner (BiRRT), which uses the reconstructed scene to generate and execute planned trajectories. Each pipeline was tested twice in each environment, resulting in 24 trials per method. For scene reconstruction, we set a maximum of $5,000$ Gaussians and used Gaussian Process interpolation length $2$ $cm$. 

\begin{figure*}[t]
    \centering
    \includegraphics[width=1.0\textwidth]{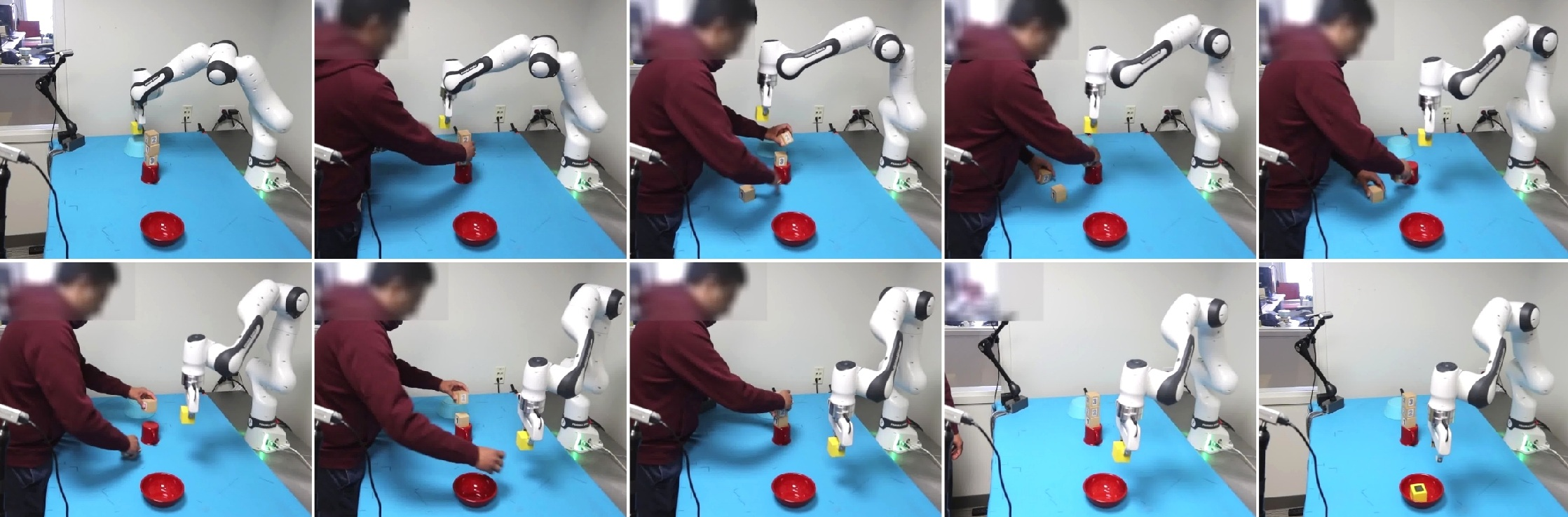}
    \caption{Sequential keyframes from the human–robot shared workspace trial. The robot transports a yellow cube from the left side of the workspace and places it into a bowl on the right, while a human operator simultaneously performs an independent task---de-stacking a disordered pile of cubes and reassembling them. Throughout the task, the robot actively avoids collisions with both the environment and the human, demonstrating the method’s effectiveness in a dynamic, shared workspace.}
    \label{fig:study_demo}
    \vspace{-0.5cm}
\end{figure*}

In our evaluation, both motion generation pipelines demonstrated high reliability with our framework, but with distinct performance characteristics. BiRRT generally produced shorter joint-space trajectories in cluttered scenarios by sampling directly in joint space (Figure~\ref{fig:real_envs_joint}). In contrast, QP-IK consistently minimized end-effector error (Figure~\ref{fig:real_envs_ee}) and produced shorter end-effector trajectories. Moreover, QP-IK maintained larger minimum distances from obstacles (Figure~\ref{fig:real_envs_min}), resulting in safer motions overall, with the exception of a single trial where it passed close to an object. Both methods largely satisfied the prescribed safety threshold (2 $cm$), though BiRRT occasionally violated it due to interpolation between sampled joint positions. Across all twelve environments, each motion generation method failed once out of the 24 trials, yielding a success rate of $95.8$\%.

\subsection{Study: Human-Robot Shared Workspace}

In addition to the preceding evaluations, we conducted a pilot study in a human–robot shared workspace under dynamic conditions. In this experiment, the robot was assigned to transport a yellow cube from the left side of the workspace and place it into a bowl on the right. At the same time, a human operator performed an independent task within the same area---de-stacking a disordered pile of cubes and reassembling them in sequence. Because the human’s pose changed continuously throughout the task, the scenario represented a highly dynamic setting with important safety considerations (see Figure \ref{fig:study_demo}). We deployed the SplatCtrl framework, incorporating both scene reconstruction and reactive motion generation. To account for the human operator, we integrated an additional Intel L515 camera (Camera 4; see Figure \ref{fig:environments}) and incorporated human segmentation~\cite{guler2018densepose} into the reconstructed scene using voxel-grid intersection. This representation was implemented similarly to the robot’s model (Section \ref{sec_method}) and accelerated using a hash-based approach. As a baseline, we implemented the industry-standard Safety Monitored Stopping (SMS) method (ISO 10218-1/2), in which the robot halts operation whenever a human enters its workspace and resumes only after the human has exited.

Eight healthy participants (six male, two female) with no prior robotic arm experience took part in the study. Each completed a single ~30-minute session and provided written informed consent. The protocol was approved by the MERL Internal IRB. Participants performed four trials: in two, the human’s cube stack was placed closer to the robot’s pickup location, and in two, closer to its goal location, with conditions randomized and counterbalanced. In all trials, obstacle shapes and positions and the human’s movements were unknown \textit{a priori} to the robot and reconstructed online with SplatCtrl. Each trial ended when both the robot and the human completed their tasks.

\begin{figure}[h]
    \centering
    \includegraphics[width=0.48\textwidth]{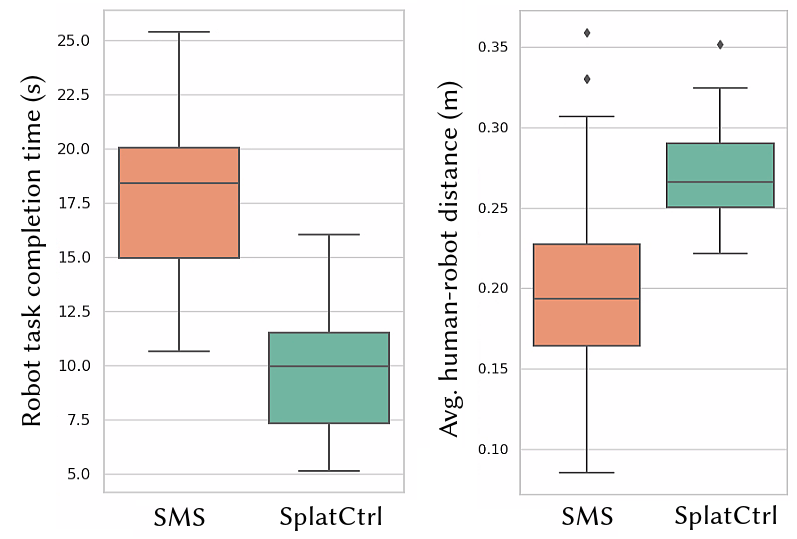}
    \caption{Compared to the Safety Monitored Stopping (SMS) baseline, SplatCtrl performs effectively in dynamic environments, achieving faster task completion times while consistently maintaining a safe distance from humans.}
    \label{fig:study_results}
    \vspace{-0.6cm}
\end{figure}

The results show that SplatCtrl safely avoided obstacles while adapting its motion to maintain separation from the human in all trials. As shown in Figure \ref{fig:study_results}, our method consistently maintained a greater average distance than the SMS baseline, demonstrating more proactive avoidance. In contrast, the SMS approach halted the robot when the human entered the workspace, often leaving it idle nearby and reducing separation. Our method also enabled faster task completion by avoiding unnecessary idle time, improving efficiency without compromising safety. These findings demonstrate its applicability in dynamic scenarios.

\section{Discussion and Conclusion}

A key implication of this work is extending 3D Gaussian Splatting beyond its traditional use in computer graphics and visualization. SplatCtrl introduces a compact, expressive, and incrementally updatable representation that integrates directly with reactive robotic control, operating without reliance on datasets or offline training. Despite these contributions, several limitations remain. First, although the method performs well in the evaluated scenarios, its scalability to large-scale environments and a wider range of manipulation tasks warrants further investigation. Second, in the current representation explicit object-level collision avoidance for the grasped item was not incorporated. Moving toward an object-centric formulation within our framework would enable more fine-grained and context-aware interactions. Third, the approach entails certain trade-offs. It requires hyperparameter tuning, and approximating Gaussians as spheres overlooks their full probabilistic characteristics. Future work will focus on computing distances directly between Gaussian distributions to preserve richer geometric and uncertainty information. We also plan to incorporate image-level spatial filtering to reduce reliance on voxel grids. Finally, as future work, the insights gained from the pilot user study on the human-robot shared workspace will be further evaluated through a larger-scale study.

In this work, we introduced SplatCtrl, a unified framework that integrates Gaussian-based scene reconstruction with reactive robot control.  The framework introduces key technical contributions for memory-efficient dynamic Gaussian Splatting,
continuous signed distance field (SDF) formulation based on Gaussian Process Distance Fields (GPDF), and reactive controller that ensures collision-free robot motion generation. Unlike prior methods, SplatCtrl operates directly from RGB-D inputs and seamlessly supports both motion planning and reactive control in unstructured, and dynamic environments. By tightly coupling perception with action, the framework reduces reliance on pre-defined maps and static assumptions. The proposed approach was evaluated through simulations, real-world robotic experiments, and a user study, demonstrating its robustness, adaptability, and practical applicability.

\bibliographystyle{IEEEtran}
\bibliography{Star} 
\end{document}